\documentclass[]{interact}

\usepackage{epstopdf}
\usepackage[caption=false]{subfig}
\usepackage{nameref}
\usepackage[longnamesfirst,sort]{natbib}
\bibpunct[, ]{(}{)}{;}{a}{,}{,}
\renewcommand\bibfont{\fontsize{10}{12}\selectfont}

\usepackage{xcolor}

\theoremstyle{plain}

\theoremstyle{definition}

\theoremstyle{remark}

\begin{document}
\articletype{Research Article}

\title{Integration of a machine learning model into a decision support tool to predict absenteeism at work of prospective employees}

\author{
\name{Gopal Nath\textsuperscript{a}\thanks{CONTACT: Gopal  Nath. Email: gnath@murraystate.edu}, Antoine Harfouche\textsuperscript{b}, Austin Coursey\textsuperscript{c}, Krishna K. Saha\textsuperscript{d}, Srikanth Prabhu\textsuperscript{e}  and Saptarshi Sengupta\textsuperscript{c}}
\affil{\textsuperscript{a}Department of Mathematics and Statistics, Murray State University, Murray, KY, USA; \textsuperscript{b}Paris-Nanterre University, Nanterre, France;  \textsuperscript{c}Department of Computer Science and Information Systems, Murray State University, Murray, KY, USA; 
	\textsuperscript{d} Department of Mathematical Sciences, Central Connecticut State University, New Britain, CT, USA;
\textsuperscript{e}Department of Computer Science and Engineering, Manipal Institute of Technology, Manipal, KA, India}
}

\maketitle

\begin{abstract}
\noindent\textbf{Purpose} - Inefficient hiring may result in lower productivity and higher training costs. Productivity losses caused by absenteeism at work cost U.S. employers billions of dollars each year. Also, employers typically spend a considerable amount of time managing employees who perform poorly. The purpose of this study is to develop a decision support tool to predict  absenteeism among potential employees.

\noindent\textbf{Design/methodology/approach} - We utilized a popular open-access dataset. In order to categorize absenteeism classes, the data have been preprocessed, and four methods of machine learning classification have been applied: Multinomial Logistic Regression (MLR), Support Vector Machines (SVM), Artificial Neural Networks (ANN), and Random Forests (RF). We selected the best model, based on several validation scores, and compared its performance against the existing model; we then integrated the best model into our proposed web-based for hiring managers.

\noindent\textbf{Findings} - A web-based decision tool allows hiring managers to make more informed decisions before hiring a potential employee, thus reducing time, financial loss and reducing the probability of economic insolvency.

\noindent\textbf{Originality/value} - In this paper, we propose a model that is trained based on attributes that can be collected during the hiring process. Furthermore, hiring managers may lack experience in machine learning or do not have the time to spend developing machine learning algorithms. Thus, we propose a web-based interactive tool that can be used without prior knowledge of machine learning algorithms.

\end{abstract}


\section*{Introduction}
 Absenteeism has been identified as an important factor in company performance and productivity losses. A study conducted by the Bureau of Labor Statistics suggests that nearly 2.8 million workdays are lost each year due to absenteeism at work \citep{Kocakulah_2016}.  Despite employers' expectations, excessive absences may reduce productivity and negatively impact the company's finances and other aspects. The Gallup Wellbeing Index surveyed over 94,000 individuals from 14 major occupations and determined that absenteeism among particular U.S. workers costs \$153 billion annually \citep{Simpson_2011}.  It is alarming to learn that many employers in the United States are unaware of the extent of absenteeism in the workplace. Less than one half of all companies have a system for tracking absenteeism, and only 16 percent have measures to reduce absenteeism  \citep{Kocakulah_2016}. Consequently, absenteeism has a significant impact on a company's financial performance since it has a direct and powerful effect on the organizational structure. In order to effectively handle employee absenteeism, an organization needs to understand the causes and patterns of absence from multiple perspectives to make a proper classification of the reasons for employee absence. In this paper, we propose a decision support tool for hiring managers that provides additional information about the potential employee based on the attributes that can be collected prior to hiring the individual. This tool enables hiring managers without expertise in the field of machine learning to gather more information and make a more informed decision about which employees to hire, thus reducing both the adverse effects of workplace absenteeism on their productivity and the company's performance. The paper is organized as follows. The proposed interactive web-based tool is described in section \nameref{literature_review_tool}, along with a brief review of the literature related to predicting work-related absences. A detailed examination of the dataset used in this study will be presented in section \nameref{Preprocessiong}. Different multiclass classification models are compared to determine the most effective model for predicting workplace absenteeism of potential candidates discussed in section \nameref{prediction_models}, along with performance metrics that are used to select the most effective model, which is later integrated with the web-based interactive tool. We describe the results obtained by the multiclass classification models and provide some recommendations for selecting the best model based on performance metrics in section \nameref{results_comparison}. By incorporating the best model found in section \nameref{results_comparison}, section \nameref{AIT} describes how to develop an absenteeism interactive tool. Moreover, we present a demonstration of how to use our proposed tool to identify the absenteeism class of a potential candidate.  Finally, some concluding remarks are provided in section \nameref{conclusion}.

\section*{Literature review and proposed method\label{literature_review_tool}}
\begin{figure}
	\begin{center}
		\includegraphics[width=6in]{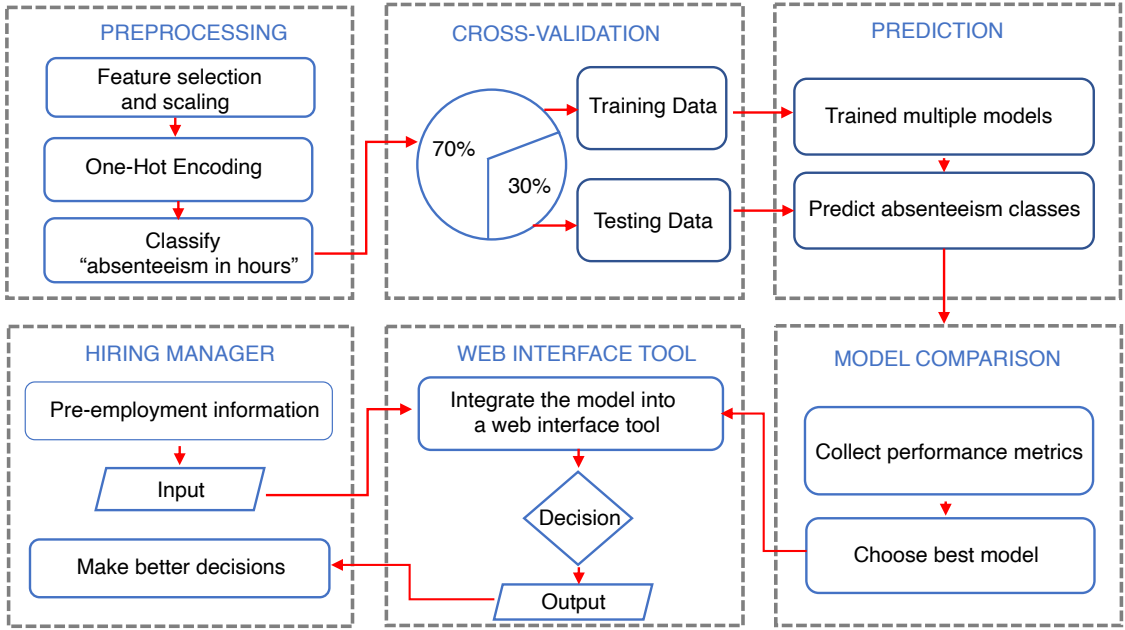}
	\end{center}
	\caption{A flowchart outlining the proposed methodology}
	\label{powerpoint_proposed_model}
\end{figure}

There are many types of employee leaves of absence, including short and long-term disability, workers' compensation, family and medical leave, and military leave. There is also evidence that companies with low morale suffer from higher rates and costs of unscheduled absences. Based on the 2005 CCH Survey, only 35\% of unscheduled absences are attributed to personal illness, but 65\% are attributed to other reasons, including family concerns (21\%), personal needs (18\%), entitlement attitude (14\%), and stress (12\%) \citep{Navarro_2006}. \citet{Tunceli_2005} explored that employees with diabetes have a 7.1\% lower likelihood of working and a 4.4\% lower likelihood of working than those without diabetes. Furthermore, the study conducted by  \citet{Halpern_2001} showed that smoking policies in the workplace affect absenteeism and productivity, concluding that smokers are more likely to be absent from work than neither former smokers nor nonsmokers \citep{Halpern_2001}. Researchers have increasingly used artificial intelligence (AI)-based methods in recent years to model absenteeism problems that have a detrimental impact on the company's infrastructure. By using Naive Bayes, Decision Trees, and Multilayer Perceptrons, Gayathri predicts absenteeism at the workplace and recommends multilayer perceptions based on the validation score  \citep{Gayathri_2018}. Using a multilayer perceptron with the error back-propagation algorithm, \citet{Martiniano_2012} proposed a neuro-fuzzy network to predict workplace absenteeism \citep{Martiniano_2012}. 
In a recent study, the authors \citep{Skorikov_2020} compared six different data mining techniques to predict absenteeism at work and concluded that KNN classifiers with the Chebyshev distance metric performed the best. Machine learning algorithms may be used to identify candidates who are more prone to being absent from work at an early stage of recruitment. Nevertheless, the above machine learning techniques are not particularly helpful to hiring managers since some attributes were gathered after recording an employee's performance. Researchers in a recent study attempted to make a model at an early stage of hiring. However, they used some attributes in their proposed model, for instance, seasons of employee absences and disciplinary failures, which information is not available at an early step of hiring. Therefore, their model cannot be beneficial to the hiring managers. Furthermore, hiring managers may not have sufficient knowledge of the data mining techniques that were used in the existing models. To resolve this problem for hiring managers, we propose an automated decision support tool that can be used to identify high probability candidates of absenteeism at an early stage in the recruitment process. For example, \citet{Delen_2007} developed an automated web-based tool integrating prediction models to provide Hollywood producers with a way to classify a movie in one of nine success categories, ranging from flops to blockbusters. \citet{Simsek_2020} have developed an automated tool that uses artificial neural networks (ANN) to identify point velocity profiles on rivers with an accuracy level of 0.46. Figure \ref{powerpoint_proposed_model} provides an overview of how the web-based interactive tool was developed and how it can be utilized. In the beginning, we will preprocess the data by performing feature selection and scaling, and One-hot encoding and classifying the absenteeism hours. We used the Synthetic Minority Over-Sampling Technique (SMOTE) to improve the performance of the classification models.  In the next step, split data into training and testing, then train four models (MLR, SVM, ANN, and RF) using training data, and predict absenteeism classes using the testing data. Utilizing performance metrics, we will compare and choose the most suitable model, and then we will integrate the selected model into a proposed web-based interactive tool. Lastly, we will provide a brief description of how the user can access the interactive tool.  

\subsection*{Preprocessing and Cross-validation \label{Preprocessiong}}

The dataset for this study was obtained from the UCI Machine Learning Repository and provided by \citet{Martiniano_2012}. This database was created using absenteeism records from a Brazilian courier company from July 2007 to July 2010. The dataset consists of 740 instances and 21 attributes. For more detailed information regarding the data set, please see \citet{Skorikov_2020}. In accordance with information provided by the data source (UCI Machine Learning Repository) and authors \citep{Martiniano_2012}, the data set permits several kinds of attributes to be combined and excluded, or type of attributes (categorical, integer, or real) can be modified depending on the purpose of research. The proposed model is to develop an interactive web-based tool for hiring managers so that, with this tool, hiring managers will be able to make better decisions based on the information provided by the candidates during the application process. In light of our proposed methodology and by utilizing expert knowledge, we will disregard the attributes \textit{Seasons, Months of absence, Days of the week,  Service time, Hit target and,  Disciplinary failure } as these attributes were unknown at the early stage of hiring. The removal of attributes without discussing their statistical importance may result in poor performance for classification problems. We will compare our model's performance with the existing model in section \nameref{results_comparison} in order to ensure that no information is lost by disregarding those attributes. An important step in the development of a machine learning model is to determine the importance of features. We applied One-Hot-Encoding (created a binary column) to the categorical attributes \citep{Imran_2019}. A number of features are redundant or do not contain useful information. Based on the importance of features, the features can be selected appropriately. We utilized the random forest classifier to determine the feature importance \citep{Saarela_2021}. We have provided a list of the top ten most significant features in Table \ref{table: feature importance}. We can see that 	\textit{Reason for absence-0}(certain infectious and parasitic diseases) has the highest importance, followed by 	\textit{Work load average/day}, 	\textit{Reason for absence-19}(external causes of morbidity and mortality),  	\textit{Reason for absence-13} (diseases of the genitourinary system), 	\textit{Distance from residence to work}, 	\textit{Age}, 	\textit{Body mass index}, 	\textit{Height},  	\textit{Weight}, and 	\textit{Transportation expense}. These can all be obtained during the hiring process. To determine the optimum F1-score, we have tested different combinations of attributes for each of the models. The combination of the attributes for which we obtain the optimum F1-score is presented in section \nameref{results_comparison}. 
\begin{table}
	\tbl{Feature importances with a Random Forest \label{table: feature importance}}
	{\begin{tabular}{lcccccc} \toprule
		
			Attribute & Feature importance  \\ \midrule
			\textit{Reason for absence-0} & 0.312  \\
			\textit{Work load average/day}	 & 0.180  \\
			\textit{Reason for absence-19}& 0.041  \\
			\textit{Reason for absence-13}& 0.041  \\
				\textit{Distance from residence to work}&  0.032\\ 
				\textit{Age} &  0.031\\ 
			 	\textit{Body mass index}  & 0.030  \\ 
			 	\textit{Height} & 0.029 \\
				\textit{Weight} &  0.028\\
			 	\textit{Transportation expense} & 0.026 \\
			 \bottomrule
	\end{tabular}}
\end{table}

As outlined in our proposed study, \textit{absenteeism in hours} will be a variable of interest. The absenteeism rate is divided into three categories: \textit{none} means that an employee is never absent, \textit{moderate} involves employees who are absent for 1 to 15 hours per month, and \textit{excessive} refers to employees who are absent for 16 to 120 hours per month. For simplicity, we refer to these three groups as $A^+$, $B^+$, and $C^+$, respectively. \citet{Skorikov_2020} introduced the concept of classification on absenteeism data, which is helpful when comparing groups within an organization. 
  Furthermore, these predefined classes provide us with the opportunity to compare our findings with existing results. 

\section*{Prediction Models\label{prediction_models}}
For our study, we analyzed four commonly used models to classify the absenteeism category of potential employees. The following subsections outline the method and the corresponding optimal hyperparameters calculated by the grid search algorithm.

\subsection*{Support vector machine} 
\begin{figure}
	\begin{center}
		\includegraphics[width=3.1in,height=2.1in]{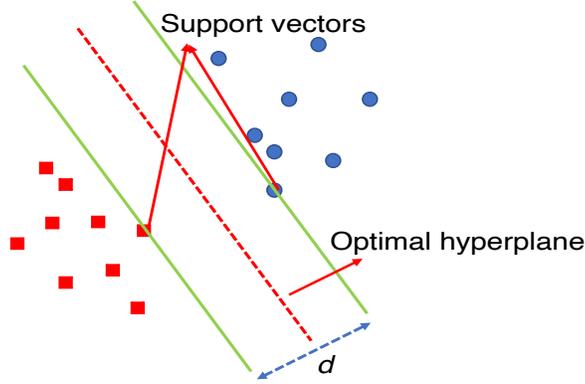}
	\end{center}
	\caption{Optimal classification algorithm}
	\label{svm}
\end{figure}

The support vector machine has become very popular because it offers significant accuracy with minimal computation power. Support vector machine (SVM) performs reasonably well with linear dependencies, have reasonable performance with sparse data sets, and can be used for a wide variety of data types \citep{Simsek_2021}. The purpose of the support vector machine algorithm is to produce a hyperplane capable of differentiating between two different classes of data. A separating hyperplane with the largest margin defined by $d=\frac{2}{||a||}$(vector $a$ is perpendicular to the separating hyperplane specifies) the maximum distance between data points of two classes as shown in Figure \ref{svm} \citep{Cervantes_2020}. The hyperplane may not be readily available in some cases due to the greater dimensions of the problem, in which case a kernel function helps in the smooth computation of the problem \citep{Vapnik_1998}. In order to apply SVM to multi-class classification problems, it is common to divide the problem into multiple binary classification subsets and then apply a standard SVM to each subset, for example the one-versus-rest technique. In order to achieve optimal accuracy, we have tested several different parameters. We found that the radial basis function kernel function provides the highest level of accuracy.

\subsection*{Multinomial logistic regression } Multinomial logistic regression is an extension of binary logistic regression in which the outcome variable can be categorized into more than two categories. In order to extend logistic regression to multiclass classification problems, one commonly used approach is to divide the multiclass classification problem into a series of binary classification sets and fit a standard logistic regression model for each subset. 
Consider $h_{mn}$   as the success ($h_{mn}=1$) or failure ($h_{mn}=0$) of multinomial outcomes $n$,~ $n=1, \hdots, N$, for observation $m,~ m=1,\hdots, M$. Consider $x_m$ denote observation $m's$ $K$-dimensional vector of the predictor variable, $k=1,\hdots, K$. 
Based on the reference outcome $N$, the multinomial logistic regression (MLR) can be defined to predict probabilities $\pi_{mn}(x_m)$ for outcomes $n=1,\hdots, N-1$ as follows \citep{Agresti_2002}.

\begin{equation}
\pi_{mn}(x_m)=\frac{\exp(\boldsymbol\lambda_n+\boldsymbol{\theta}'_n\boldsymbol{x}_m)}{1+\sum_{k=1}^{J-1} \exp(\boldsymbol\lambda_k+\boldsymbol{\theta}'_k\boldsymbol{x}_m)},
\end{equation}
\noindent where $\boldsymbol{\theta}_n= (\theta_{n1},\hdots,\theta_{nK})'$ refers the  coefficients for the $n$th linear predictor are given excluding its intercept $\lambda_n$.  
The log-likelihood method is used to estimate  $\boldsymbol\lambda$ and $\boldsymbol{\theta}$, providing normal and consistent estimates. The Newton method algorithm was used to optimize the problem for the most accurate prediction.

\subsection*{Artificial neural networks} 
\begin{figure}
	\centering
	\subfloat[The input values are transformed in a perceptron by the weights, biases, and activation functions. Output values are sent to the next perceptron.]{%
		\resizebox*{7cm}{!}{\includegraphics{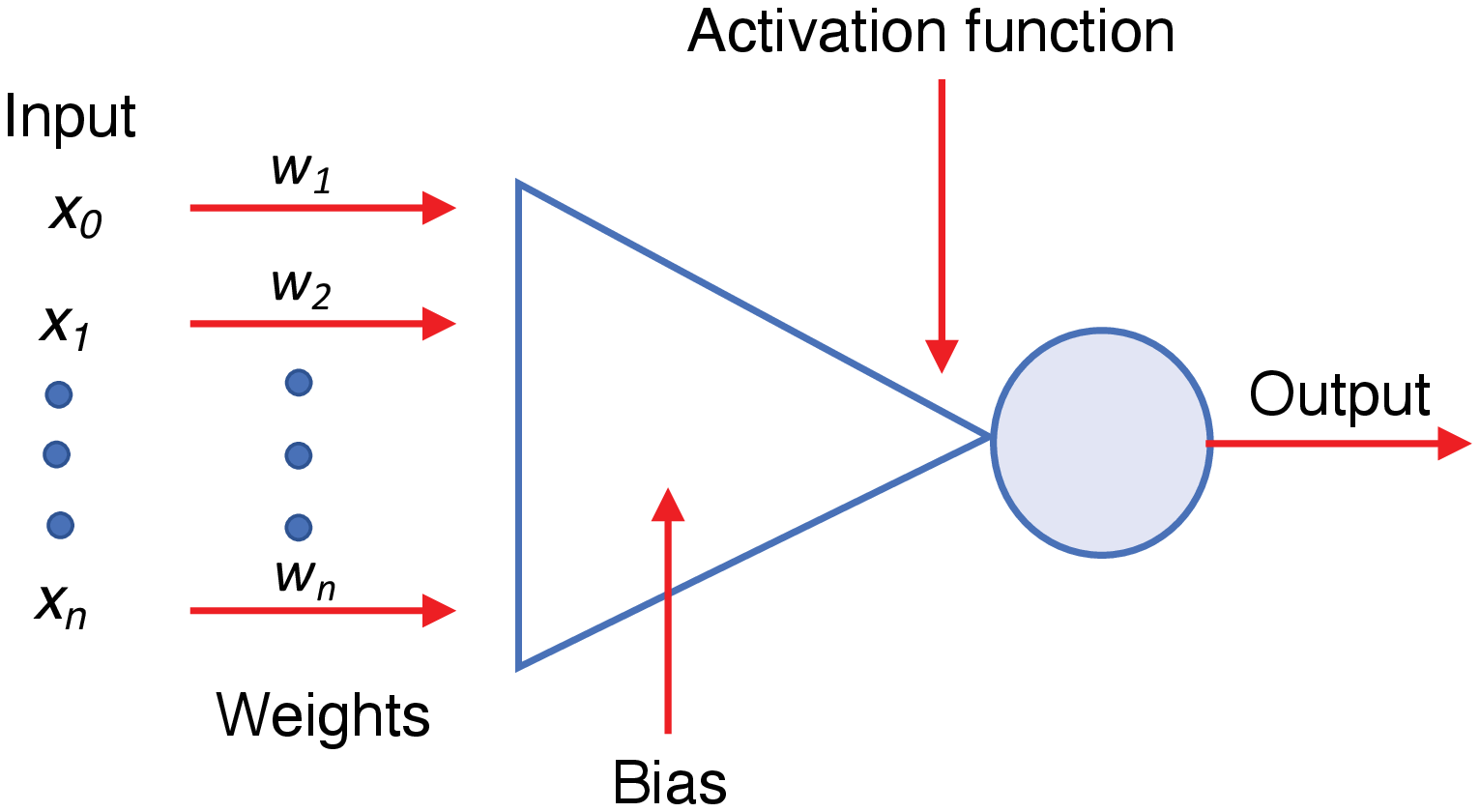}}}\hspace{5pt}
	\subfloat[Multilayered perceptron is composed of several perceptrons.]{%
		\resizebox*{6cm}{!}{\includegraphics{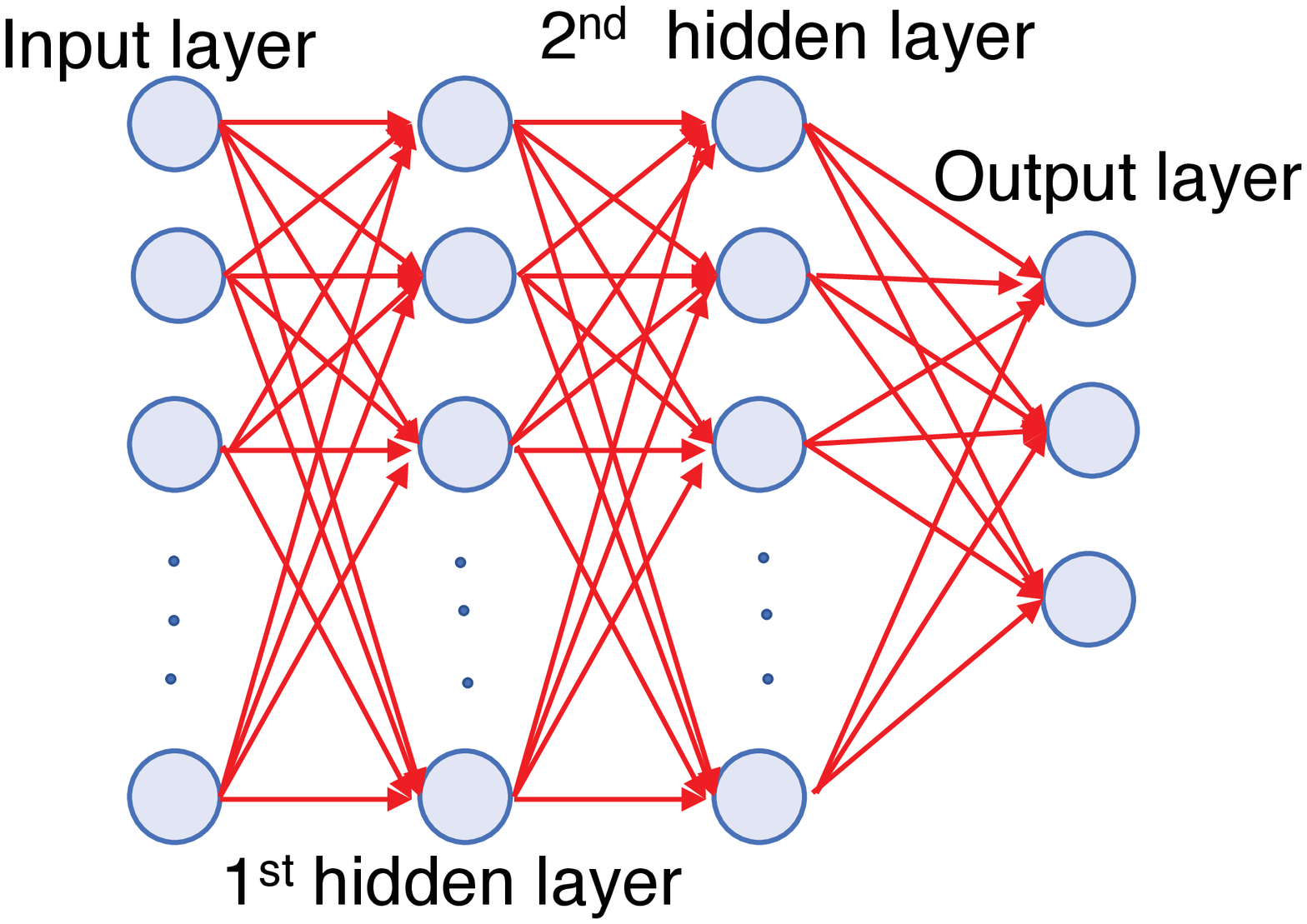}}}
	\caption{Architecture of artificial neural networks} \label{architecture_ann}
\end{figure}

In recent years, artificial neural network (ANN) has been applied to various fields due to its ability to model highly challenging problems. A new and useful model is found to be the ANN when applied to problem-solving and machine learning. This is a model of information management that is comparable to the function of the human nervous system. A key feature of this brain is its ability to process information in a unique manner. Many interconnected \textit{neurons} serve as components of the system that work in concert to solve specific problems on a daily basis \citep{Abiodun_2018}. An ANN consists of nodes, representing \textit{neurons}, and connections between \textit{nodes}, representing axons and dendrites carrying information. There is a value or weight attached to every connection between two nodes for the purpose of assessing the strength of the signal \citep{Yadav_2018}. The \textit{neurons} are arranged in layers, with an input layer representing one type of input data, an output layer representing the result of the classification, and one or more hidden layers. One of the most common and widely used forms of ANN is the perceptron, which is a fully connected feed-forward network \citep{Hallinan_2013}. The linear combination of weights and input values is passed through a non-linear function known as an activation function \citep{Young_2014}. Neural activation functions approximate the complex physical processes of neurons, which modulate their output in a non-linear way.  The architecture of an artificial neural network is shown in Figure \ref{architecture_ann} \citep{Kim_2018}. 
We built a six-layer fully connected ANN, in which each neuron in one layer is connected to every neuron in the following layer. After one hot encoding and standardization, the first input layer consists of 42 input neurons. Our results indicate that the highest degree of accuracy will be achieved by using a network with four hidden layers consisting of 400 -100 -50 -20 neurons(nodes) to the output layer consisting of three neurons using the \textit{relu} activation function.

\subsection*{Random forest } Decision trees are the core component of random forest classifiers. Using the features of a data set, a decision tree is built using a hierarchical structure. In the decision tree, each node represents a measure associated with a subset of features \citep{Suthaharan_2016}. A random forest is composed of trees that produce class predictions for each tree, and the class that receives the most votes becomes the model prediction \citep{Sarica_2017}. In this study, Gini impurity supported criteria were used to determine most accurate predictions.

\subsection*{Performance Metrics} 
\noindent The following section demonstrates some of the statistical assessment metrics that we used to validate our model's performance.   

\noindent \textbf{Accuracy:} An important metric for evaluating classification models is accuracy.  Accuracy can also be determined based on binary classification as follows: 
\begin{equation}
\text{Accuracy}= \frac{TP+TN}{TP+TN+FP+FN},
\end{equation}

\noindent where $TP$ stands for true positives, $TN$ for true negatives, $FP$ for false positives, and $FN$ stands for false negatives \citep{Kulkarni_2020}. 

With imbalanced datasets, accuracy may be misleading; therefore, there are additional metrics based on confusion matrix that can be utilized to evaluate performance.

\noindent \textbf{Precision:} Precision is a popular metric in classification systems. A measure of how well a model is able to predict positive values, referred to as precision \citep{Kulkarni_2020}: 
\begin{equation}
\text{Precision} =\frac{TP}{TP+FP}
\end{equation}

\noindent\textbf{Recall:} Recall  addresses imbalances that may occur in a dataset, is defined as follows \citep{Leonard_2017}:
\begin{equation}
\text{Recall}= \frac{TP}{TP+FN},
\end{equation}

\noindent \textbf{F1-score:} In terms of precision and recall, the F-measure is defined as follows \citep{Sun_2020}: 
\begin{equation}
F-\text{score} = (1+ \alpha^2 ) \frac{P \times R}{\alpha^2 (P+R) },
\end{equation}

 \noindent where  \textit{precision} and \textit{recall} are represented by $P$ and $R$ respectively,  and   $\alpha \ge 0$ represents the balance between $P$ and $R$. It is commonly referred to as F1 score when $\alpha = 1$ \citep{Zhang_2021}.

\noindent \textbf{ROC AUC score}: Over the past few decades, receiver operating characteristic (ROC) curves have become the most popular and have been used as a tool to evaluate the discrimination ability of various machine learning methods for predictive purposes \citep{Hanley_1982}. Better models will pass through the upper left corner and have greater overall testing accuracy. One of the most widely used metrics for assessing the performance of models is the Area Under Curve (AUC). It provides the ability for the classifier to distinguish between classes and is used as a summary of the ROC curve. AUC values are generally between 0.5 and 1.0, and a larger AUC indicates better performance. In the case of a perfect model, its AUC would be 1, indicating that all positive examples are always in front of all negative examples \citep{Zhang_2021}. There are two popular methods for evaluating multi-class classification problems. The one-vs-one algorithm computes the average of the pairwise ROC AUC scores, while the one-vs-rest algorithm computes the average of the ROC AUC scores for each class compared to all the other classes \citep{Fawcett_2006}.

\section*{Results and   Model Comparison\label{results_comparison}} In this article, we explored four popular machine learning algorithms for predicting employee absenteeism. We have tested various training and test data sets ratios, and based on the best F1-score, we trained each model using 70\% of the data and then evaluated its efficiency using the remaining 30\%. We subsequently selected the most suitable model out of four using the performance metrics and developed a decision support tool for hiring managers. To obtain maximum accuracy we tested different settings for each of the methods. The 42 attributes were used (after performing the One-hot encoding technique), the quantative attributes were normalized and SMOTE was applied to the training set to train SVM, ANN, and RF models. Our selection of attributes for MLR based on the best performance metrics are \textit{Transportation expense, Work load average/day(normalized), Distance from residence to work, Age, Son, Social drinker, Social smoker, Pet, Weight, Body mass index, Reason for absence}, and \textit{Education}. Having categorized employees' absenteeism in hours, which generates imbalance classes. There have been numerous suggestions made to overcome the negative effects of an imbalance on performance metrics, for further details please see \citet{Luque_2019}. On the basis of our experimental results and  suggested by \citet{Johnson_2019} and \citet{Simsek_2020}, we determined the best model based on the F1-score and confusion matrix. 
In Table \ref{table:performance_metirics}, we reported accuracy, weighted F1-score, weighted precision, and the one-vs-one ROC AUC scores weighted by prevalence. Without oversampling, the MLR achieved the highest accuracy, precision, recall, F1-score, and ROC AUC score out of all algorithms. \citet{Skorikov_2020} proposed a KNN classifier with a Chebyshev distance metric that achieved accuracy, precision, recall, and F1-scores of respectively 0.923, 0.83, 0.91, and 0.87.  By comparing the results in Table \ref{table:performance_metirics}, it can be seen that the MLR model performs better than their proposed model under all scenarios. A major reason why we chose MLR for the web-based supporting tool is that accuracy for class $A^+$ is 91\% (9\% incorrectly classified as $B^+$), for class $B^+$ is 100\%, and for class $C^+$ is 27\%. As it can be seen that class $A^+$and class $B^+$ do not get misclassified as class $C^+$, the proposed tool should be on the safe side in order to prevent removing any good candidates during the process of filtering.

\begin{table}
	\tbl{Performance of the trained  models when applied to the test data. \label{table:performance_metirics}}
	{\begin{tabular}{lcccccc} \toprule
			
			Model & Accuracy & Precision    & Recall  & F1-score& ROC AUC score  \\ \midrule
		MLR & 0.932            & 0.937           &0.932  & 0.915   & 0.885 \\
		 	 SVM &0.887        &  0.916      &  0.887    & 0.898 &0.870 \\
		 	ANN& 0.873        &0.897      & 0.873  & 0.884  & 0.874 \\
	RF &0.869                &0.880           & 0.869  &0.874& 0.838\\
	
			\bottomrule
	\end{tabular}}
\end{table}


\section*{Absenteeism interactive tool\label{AIT}} 
\begin{figure}
	\begin{center}
		\includegraphics[width=5.65in]{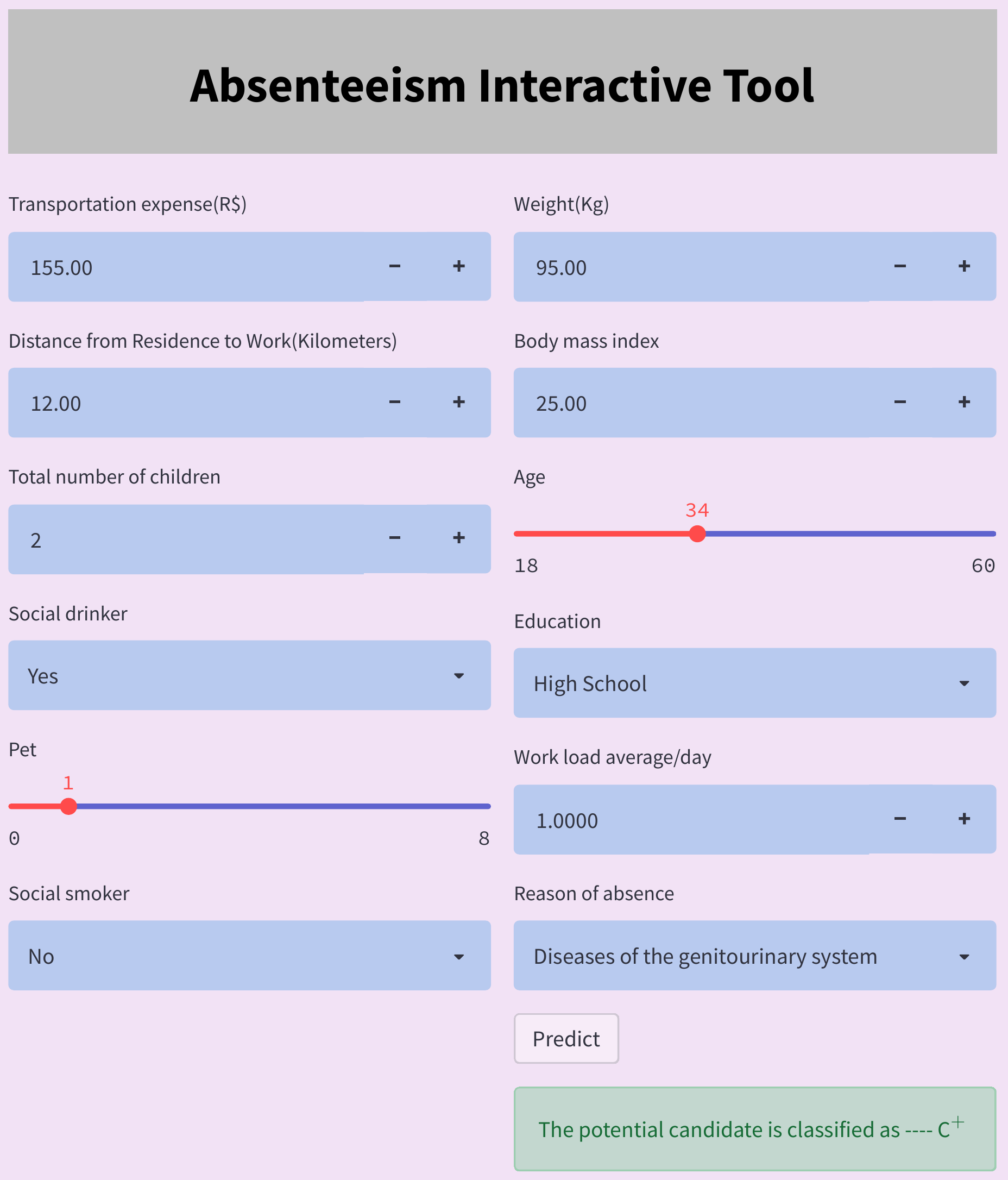}
	\end{center}
	\caption{A screenshot of the proposed interactive tool. Based on input values, the tool classifies the candidate as $C^+$ on a scale of  $A^+B^+C^+$.}
	\label{C}
\end{figure}

It can be difficult for a hiring manager to predict employee absenteeism classes using a machine learning algorithm, especially if he or she lacks programming expertise. Our proposed web-based interactive tool would enable hiring managers to make better decisions at the early stages of hiring by filtering potential applicants. The hiring manager will save a significant amount of time by not manually screening all applicants.  In this article, we present a prototype of the proposed tools based on an open-access dataset, which is publicly available and was collected by \citet{Martiniano_2012}  from  a Brazilian courier company in 2012. The dataset is widely used and there has been a significant amount of research published on absenteeism using this dataset. There are features in the dataset that reference human behavior, which may correspond differently in other areas of the world \citep{Shah_2020}. An adaptation of the model can be made to other global locations by adapting it to the local characteristics of the employees.  In order to protect individual traits based on geographical location, hiring managers or human resources should carefully select the attributes.  Based on the performance analysis discussed in the previous section, MLR was applied to the proposed web-based tool. We developed the proposed tool using \textit{Streamlit}, an open-source Python framework for developing web applications. By collecting all the information of a potential job candidate, the end-user can then input all of the information and, by clicking on the predict button, can predict the class of the candidate. As shown in Figure \ref{C}, the output(predicted) class for the desired candidate is $C ^+$on a  scale  of $A^+B^+C^+$, which means that this candidate is more likely to have more absences than other candidates.

\newpage 
\section*{Conclusions\label{conclusion}} The cost of absenteeism is billions of dollars and the time spent managing employees who are not performing well. Thus, absenteeism has a negative effect on a company's organizational structure. Companies differ depending on their working environment, type of work, geographical location, etc. Therefore, in order to effectively handle employee absenteeism, which adversely affects a company's financial stability, the company needs to identify the cause of the problem. Consequently, the company spent large amounts of money hiring additional management staff to identify the cause and find a suitable solution. As a result, to avoid any negative impacts on the company, the machine learning algorithm can determine the potential class of such employees in terms of punctuality of the work at the initial stage of hiring. Furthermore, using the machine learning algorithm can be quite challenging since it requires a high level of knowledge, a significant amount of time, and strong programming abilities. The hiring manager may not have enough experience with machine learning algorithms or may not have the time to develop such an algorithm. To address all these issues, we studied an open access data set, applied feature selection, scalability and cross validation, and then trained multiple models (MLR, SVM, ANN and RF). In comparison of their performance using different performance metrics, MLR performs best in terms of F1-score and confusion matrix. Furthermore, we compare the performance of MLR with the existing proposed models, and MLR outperforms all of them. We proposed MLR  to predict absenteeism at work, which is an easy method to explain, and contains no so-called black box. In the next step, we utilized the Python Streamlit framework to build a web-based interactive tool by integrating the MLR model. This web-based tool serves as a link between machine learning algorithms and hiring managers. 
Finally, our proposed interactive tool plays a crucial role in solving those consequences by simply utilizing it at an early stage in the hiring process. End-users don't need to have any previous knowledge of how machine learning algorithms work to use our proposed tools. By utilizing this simple tool, hiring managers can save time, reduce the workload of the hiring process, and make better hiring decisions, thereby preserving the company from adverse financial consequences.

\end{document}